\documentclass[11pt]{ifacconf}

\usepackage{natbib}            
\usepackage{graphicx}          
\usepackage{graphicx}         
\usepackage{amsfonts}
\usepackage{amsmath}
\usepackage{psfrag}

\newcommand{\ie}{{\it i.e.}}
\newcommand{\argmin}{\mathop{\rm argmin}}

\newcommand{\symm}{{\mbox{\bf S}}}
\newcommand{\Tr}{\mathop{\bf Tr}}
\newcommand{\diag}{\mathop{\bf diag}}

\newcommand{\reals}{\mathbf R}

\begin{document}

\begin{frontmatter}
\title{An ADMM Algorithm for a Class of Total Variation Regularized Estimation
Problems\thanksref{footnoteinfo}} 

\thanks[footnoteinfo]{This work was partially supported by the Swedish Research
Council, the Linnaeus Center ACCESS at KTH and the European Research Council
under the advanced grant LEARN, contract 267381.  }

\author[First]{Bo Wahlberg},
\author[Second]{Stephen Boyd},
\author[First]{Mariette Annergren},
\author[Second]{ and Yang Wang}

\address[First]{Automatic Control Lab and ACCESS, School of Electrical
Engineering, KTH Royal Institute of Technology, \\ SE 100 44 Stockholm,Sweden}
\address[Second]{Department of Electrical Engineering, Stanford University,
Stanford, CA 94305, USA}

\begin{keyword}                           
Signal processing algorithms, stochastic  parameters, parameter estimation,
convex optimization and regularization
\end{keyword}                             

\begin{abstract}                          
We present an alternating augmented Lagrangian method for convex optimization
problems where the cost function is the sum of two terms, one that is
separable in the variable blocks, and a second that is separable in the
difference between consecutive variable blocks. Examples of such problems
include Fused Lasso estimation, total variation denoising, and
multi-period portfolio optimization with transaction costs.
In each iteration of our
method, the first step involves separately optimizing over each variable block,
which can be carried out in parallel.
The second step is not separable in the
variables, but can be carried out very efficiently.
We apply the algorithm to segmentation of data based on changes in
mean ($\ell_1$ mean filtering) or changes
in variance ($\ell_1$ variance filtering). In a numerical example, we show
that our implementation is around 10000 times faster compared with the generic
optimization solver SDPT3.
\end{abstract}

\end{frontmatter}

\section{Introduction}


In this paper we consider optimization problems where the objective is a sum of
two terms: The first term is separable in the variable blocks, and the second
term is separable in the difference between consecutive variable blocks. One
example is the Fused Lasso method in statistical learning,
\cite{Tibshirani-Saunders-Rosset-Zhu-Knight-05}, where the objective includes
an $\ell_1$-norm penalty on the parameters, as well as an $\ell_1$-norm
penalty on the difference between consecutive parameters. The first penalty
encourages a sparse solution, \ie, one with few nonzero entries, while the
second penalty enhances block partitions in the parameter space.
The same ideas have been applied in many other areas, such as Total Variation
(TV) denoising, \cite{Rudin:1992:NTV:142273.142312}, and segmentation of
ARX models, \cite{OhlssonLB:10}
(where it is called sum-of-norms regularization).
Another example is multi-period portfolio optimization, where the
variable blocks give the portfolio in different time periods, the first
term is the portfolio objective (such as risk-adjusted return), and
the second term accounts for transaction costs.

In many applications, the optimization problem involves a large number of
variables, and cannot be efficiently handled by generic optimization solvers.
In this paper, our main contribution is to derive an efficient and scalable
optimization algorithm, by exploiting the structure of the optimization
problem.  To do this, we use a distributed optimization method called
Alternating Direction Method of Multipliers (ADMM).  ADMM was developed in the
1970s,  and is closely related to many other optimization algorithms
including Bregman iterative algorithms for $\ell_1$ problems, Douglas-Rachford
splitting, and proximal point methods; see \cite{Eckstein92onthe, 4407760}.
ADMM has been applied in many areas, including image and signal processing,
\cite{DBLP:journals/ijcv/Setzer11}, as well as large-scale problems in
statistics and machine learning, \cite{DBLP:journals/ftml/BoydPCPE11}.


We will apply ADMM to $\ell_1$ mean filtering and $\ell_1$ variance filtering
(\cite{BW_Asilomar}), which are important problems in signal processing with
many applications, for example in financial or biological data analysis.  In
some applications, mean and variance filtering are used to pre-process data
before fitting a parametric model. For non-stationary data it is also important
for segmenting the data into stationary subsets. The approach we present is
inspired by the $\ell_1$ trend filtering method described in
\cite{Kim-Koh-Boyd-Gorinevsky-09}, which tracks changes in the mean value of
the data. (An example in this paper also tracks changes in the variance of the
underlying stochastic process.) These problems are closely related to the
covariance selection problem, \cite{Dempster-72}, which is a convex
optimization problem when the inverse covariance is used as the optimization
variable, \cite{Banerjee-ElGhaoui-dAspremont-08}.  The same ideas can also be
found in \cite{Kim-Koh-Boyd-Gorinevsky-09} and
\cite{Friedman-Hastie-Tibshirani-08}.

This paper is organized as follows. In Section  \ref{sec:admm} we review the
ADMM method. In Section \ref{sec:sep}, we apply ADMM to our optimization
problem to derive an efficient optimization algorithm.  In Section
\ref{sec:l1mean} we apply our method to $\ell_1$ mean filtering, while in
Section \ref{sec:l1var} we consider $\ell_1$ variance filtering.
Section \ref{sec:num}
contains some numerical examples, and Section \ref{sec:con} concludes the
paper.

\section{Alternating Direction Method of Multipliers (ADMM) }
\label{sec:admm}
In this section we give an overview of ADMM. We follow closely the development in Section 5 of
\cite{DBLP:journals/ftml/BoydPCPE11}.

Consider the following optimization problem
\begin{equation}\label{e-constrained-problem}
\begin{array}{ll}
\mbox{minimize} & f(x)\\
\mbox{subject to} &  x \in {\mathcal C}
\end{array}
\end{equation}
with variable $x\in  \mathbb{R}^n$, and where $f$ and $\mathcal{C}$ are convex.
We let $p^\star$ denote the optimal value of (\ref{e-constrained-problem}).
We first re-write the problem as
\begin{equation}\label{e-admm-problem}
\begin{array}{ll}
\mbox{minimize} & f(x) + I_\mathcal{C}(z)\\
\mbox{subject to} &  x = z,
\end{array}
\end{equation}
where $I_\mathcal{C}(z)$ is the indicator function on $\mathcal{C}$ (\ie,
$I_\mathcal{C}(z) = 0$ for $z\in\mathcal{C}$, and $I_\mathcal{C}(z) = \infty$
for $z\notin\mathcal{C}$).  The augmented Lagrangian for this problem is
\[
L_\rho (x, z, u) = f(x) + I_\mathcal{C}(z) + (\rho/2)\|x-z+u\|_2^2,
\]
where $u$ is a scaled dual variable associated with the constraint $x = z$,
\ie, $u = (1/\rho)y$, where $y$ is the dual variable for $x = z$.
Here, $\rho > 0$ is a penalty parameter.

In each iteration of ADMM, we perform alternating minimization of the augmented
Lagrangian over $x$ and $z$. At iteration $k$ we carry out the following steps
\begin{align}
x^{k+1} &:= \argmin_x\{f(x) +(\rho/2)\|x - z^k +
u^k\|_2^2\} \label{eq:admm1}\\
z^{k+1} &:= \Pi_{\mathcal{C}}(x^{k+1} + u^k) \label{eq:admm2}\\
u^{k+1} &:= u^k + (x^{k+1} - z^{k+1}) \label{eq:admm3},
\end{align}
where $\Pi_{\mathcal C}$ denotes Euclidean projection onto $\mathcal{C}$.
In the first step of ADMM, we fix $z$
and $u$ and minimize the augmented Lagrangian over $x$; next, we fix $x$ and
$u$ and minimize over $z$; finally, we update the dual variable $u$.

\subsection{Convergence}
Under mild assumptions on $f$ and $\mathcal{C}$, we can show that the
iterates of ADMM converge to a solution; specifically, we have
\[
f(x^k) \rightarrow p^\star, \quad x^k-z^k\rightarrow 0,
\]
as $k\rightarrow\infty$.  The rate of convergence, and hence the number of
iterations required to achieve a specified accuracy, can depend strongly on the
choice of the parameter $\rho$. When $\rho$ is well chosen, this method can
converge to a fairly accurate solution (good enough for many applications),
within a few tens of iterations.  However, if the choice of $\rho$ is poor,
many iterations can be needed for convergence.  These
issues, including heuristics for choosing $\rho$, are discussed in more detail
in \cite{DBLP:journals/ftml/BoydPCPE11}.

\subsection{Stopping criterion}
The primal and dual residuals at iteration $k$ are given by
\[
e_p^k = (x^k-z^k), \quad
e_d^k = -\rho (z^k-z^{k-1}).
\]
We terminate the algorithm when the primal and dual residuals satisfy a
stopping criterion (which can vary depending on the requirements of the
application). A typical criterion is to stop when
\[
\|e_p^k\|_2 \leq \epsilon^\mathrm{pri},\quad
\|e_d^k\|_2 \leq \epsilon^\mathrm{dual}.
\]
Here, the tolerances
$\epsilon^\mathrm{pri} > 0$ and $\epsilon^\mathrm{dual} > 0$ can
be set via an absolute plus relative criterion,
\begin{align*}
&\epsilon^\mathrm{pri} =
\sqrt{n} \epsilon^\mathrm{abs} + \epsilon^\mathrm{rel}
\max\{\|x^k\|_2, \|z^k\|_2\}, \\
&\epsilon^\mathrm{dual} =
\sqrt{n} \epsilon^\mathrm{abs} + \epsilon^\mathrm{rel}
\rho \|u^k\|_2,
\end{align*}
where $\epsilon^\mathrm{abs} > 0$ and $\epsilon^\mathrm{rel} > 0$ are absolute
and relative tolerances (see \cite{DBLP:journals/ftml/BoydPCPE11} for details).

\section{Problem formulation and method}
\label{sec:sep}
In this section we formulate our problem and derive an efficient distributed
optimization algorithm via ADMM.

\subsection{Optimization problem}
We consider the problem
\begin{equation}\label{e-our-problem}
\begin{array}{ll}
\mbox{minimize} & \sum_{i=1}^N \Phi_i(x_i)+\sum_{i=1}^{N-1} \Psi_i(r_i)\\
\mbox{subject to} & r_i=x_{i+1}-x_i,\quad i = 1,\ldots,N-1
\end{array}
\end{equation}
with variables $x_1, \ldots, x_N,r_1, \ldots, r_{N-1}\in\reals^n$,
and where $\Phi_i:\reals^n\rightarrow\reals\cup\{\infty\}$ and
$\Psi_i:\reals^n\rightarrow\reals\cup\{\infty\}$ are convex functions.

This problem has the form (\ref{e-constrained-problem}), with variables
$x = (x_1,\ldots,x_N)$, $r = (r_1,\ldots,r_{N-1})$, objective function
\[
f(x,r) = \sum_{i=1}^N \Phi_i(x_i)+\sum_{i=1}^{N-1} \Psi_i(r_i)
\]
and constraint set
\begin{equation}\label{e-constraint-set}
\mathcal{C} = \{ (x, r) \mid r_i = x_{i+1}-x_i, \;i=1,\ldots,N-1\}.
\end{equation}

The ADMM form for problem (\ref{e-our-problem}) is
\begin{equation}\label{e-our-problem-admm}
\begin{array}{ll}
\mbox{minimize} & \sum_{i=1}^N \Phi_i(x_i)+\sum_{i=1}^{N-1} \Psi_i(r_i) +
I_\mathcal{C}(z,s) \\
\mbox{subject to}
& r_i = s_i, \quad i = 1,\ldots,N-1 \\
& x_i = z_i, \quad i = 1,\ldots,N,
\end{array}
\end{equation}
with variables $x = (x_1,\ldots,x_N)$, $r = (r_1,\ldots,r_{N-1})$, $z
= (z_1,\ldots,z_N)$, and $s = (s_1,\ldots,s_{N-1})$. Furthermore, we let $u =
(u_1,\ldots,u_N)$ and $t = (t_1,\ldots,t_{N-1})$ be vectors of scaled dual
variables associated with the constraints $x_i = z_i$, $i = 1,\ldots,N$, and
$r_i = s_i$, $i = 1,\ldots,N-1$ (\ie, $u_i = (1/\rho)y_i$, where $y_i$ is the
dual variable associated with $x_i = z_i$).


\subsection{Distributed optimization method}

Applying ADMM to problem (\ref{e-our-problem-admm}), we carry out the following
steps in each iteration.

\paragraph*{Step 1.}
Since the objective function $f$ is separable in $x_i$ and $r_i$, the
first step (\ref{eq:admm1}) of the ADMM algorithm consists of $2N-1$
separate minimizations
\begin{equation}\label{e-admm-11}
x_i^{k+1} :=
\argmin_{x_i} \{\Phi_i(x_i) +(\rho/2)\|x_i - z_i^k + u_i^k\|_2^2\},
\end{equation}
$i = 1,\ldots,N$, and
\begin{equation}\label{e-admm-12}
r_i^{k+1} :=
\argmin_{r_i} \{\Psi_i(r_i) +(\rho/2)\|r_i - s_i^k + t_i^k\|_2^2\},
\end{equation}
$i = 1,\ldots,N-1$. These updates can all be carried out in parallel. For
many applications, we will see that we can often solve (\ref{e-admm-11})
and (\ref{e-admm-12}) analytically.

\paragraph*{Step 2.}
In the second step of ADMM, we project $(x^{k+1} + u^k, r^{k+1} + t^k)$ onto the constraint
set $\mathcal{C}$, \ie,
\[
(z^{k+1}, s^{k+1}) := \Pi_\mathcal{C}((x^{k+1}, r^{k+1}) + (u^k, t^k)).
\]
For the particular constraint set (\ref{e-constraint-set}), we will show in
Section \ref{s-projection} that the projection can be performed extremely
efficiently.

\paragraph*{Step 3.}
Finally, we update the dual variables:
\[
u_i^{k+1} := u_i^k + (x_i^{k+1}-z_i^{k+1}), \quad i = 1,\ldots,N
\]
and
\[
t_i^{k+1} := t_i^k + (r_i^{k+1}-s_i^{k+1}), \quad i = 1,\ldots,N-1.
\]
These updates can also be carried out independently in parallel, for each
variable block.

\subsection{Projection}\label{s-projection}

In this section we work out an efficient formula for projection onto the
constraint set $\mathcal{C}$ (\ref{e-constraint-set}).  To perform the
projection
\[
(z, s) = \Pi_\mathcal{C}((w, v)),
\]
we solve the optimization problem
\[
\begin{array}{ll}
\mbox{minimize} & \|z - w\|_2^2 + \|s - v\|_2^2 \\
\mbox{subject to} & s = Dz,
\end{array}
\]
with variables $z = (z_1,\ldots,z_N)$ and $s = (s_1,\ldots,s_{N-1})$, and where
$D\in\reals^{(N-1)n\times Nn}$ is the forward difference operator, \ie,
\[
D = \left[
\begin{array}{lllll}
-I & I &  &  &  \\
 & -I & I &  &  \\
 &  & \ddots & \ddots & \\
 &  &  & -I & I \\
\end{array}
\right].
\]
This problem is equivalent to
\[
\begin{array}{ll}
\mbox{minimize} & \|z - w\|_2^2 + \|Dz - v\|_2^2.
\end{array}
\]
with variable $z = (z_1,\ldots,z_N)$. Thus to perform the projection we first
solve the optimality condition
\begin{equation}\label{e-opt-cond}
(I + D^TD)z = w + D^Tv,
\end{equation}
for $z$, then we let $s = Dz$.

The matrix $I + D^TD$ is block tridiagonal, with diagonal blocks equal to
multiples of $I$, and sub/super-diagonal blocks equal to $-I$. Let $LL^T$ be
the Cholesky factorization of $I + D^TD$. It is easy to show that $L$ is block
banded with the form
\[
L = \left[
\begin{array}{lllll}
l_{1,1} &  &  &  & \\
l_{2,1} & l_{2,2} &  &  & \\
 & l_{3,2} & l_{3,3} &  & \\
 &  &  \ddots  & \ddots & \\
 &  &  &  l_{N,N-1} & l_{N,N}
\end{array}
\right] \otimes I,
\]
where $\otimes$ denotes the Kronecker product. The coefficients $l_{i,j}$ can
be explicitly computed via the recursion
\[
\begin{array}{l}
l_{1,1} = \sqrt{2}, \\
l_{i+1,i} = -1/l_{i,i}, \;\; l_{i+1,i+1} = \sqrt{3-l_{i+1,i}^2},
\;\; i = 1,\ldots,N-2, \\
l_{N,N-1} = -1/l_{N-1,N-1}$, \quad $l_{N,N} = \sqrt{2-l_{N,N-1}^2}.
\end{array}
\]
The coefficients only need to be computed once, before the projection
operator is applied.

The projection therefore consists of the following steps
\begin{enumerate}
\item Form $b := w + D^Tv$:
\[
\begin{array}{l}
b_1 := w_1 - v_1, \quad b_N := w_N + v_{N-1}, \\
b_i := w_i + (v_{i-1} - v_i),\quad i = 2,\ldots,N-1.
\end{array}
\]
\item Solve $Ly = b$:
\begin{align*}
y_1 &:= (1/l_{1,1})b_1, \\
y_i &:= (1/l_{i,i})(b_i - l_{i,i-1}y_{i-1}),\quad i = 2,\ldots,N.
\end{align*}
\item Solve $L^Tz = y$:
\begin{align*}
z_N &:= (1/l_{N,N})y_N, \\
z_i &:= (1/l_{i,i})(y_i-l_{i+1,i}z_{i+1}),\quad i = N-1,\ldots,1.
\end{align*}
\item Set $s = Dz$:
\[
s_i := z_{i+1}-z_i,\quad i = 1,\ldots,N-1.
\]
\end{enumerate}
Thus, we see that we can perform the projection very efficiently,
in $\mathcal{O}(Nn)$ flops (floating-point operations). In fact,
if we pre-compute the inverses $1/l_{i,i}$, $i = 1,\ldots,N$, the only
operations that are required are multiplication, addition, and
subtraction. We do not need to perform division, which can be expensive
on some hardware platforms.

\section{Examples}
\subsection{$\ell_1$ Mean filtering}
\label{sec:l1mean}

Consider a sequence of vector random variables
\[
Y_i\sim {\mathcal{N}}(\bar y_i, \Sigma),
\quad i = 1,\ldots,N,
\]
where $\bar y_i\in\reals^n$ is the mean, and $\Sigma\in\symm^n_+$ is the
covariance matrix. We assume that the covariance matrix is known, but the mean
of the process is unknown.  Given a sequence of observations $y_1,\ldots,y_N$,
our goal is to estimate the mean under the assumption that it is piecewise
constant, \ie, $\bar y_{i+1} = \bar y_i$ for many values of $i$.

In the Fused Group Lasso method, we obtain our estimates by solving
\[
\begin{array}{ll}
\mbox{minimize} & \sum_{i=1}^N\frac 1 2
(y_i-x_i)^T\Sigma^{-1}(y_i-x_i)+\lambda \sum_{i=1}^{N-1}
\|r_i\|_2\\
\mbox{subject to} & r_i=x_{i+1}-x_i, \quad i = 1,\ldots,N-1,
\end{array}
\]
with variables $x_1,\ldots,x_N$, $r_1,\ldots,r_{N-1}$. Let
$x_1^\star,\ldots,x_N^\star$, $r_1^\star,\ldots,r_{N-1}^\star$
denote an optimal point, our estimates
of $\bar y_1,\ldots,\bar y_N$ are $x_1^\star,\ldots,x_N^\star$.

This problem is clearly in
the form (\ref{e-our-problem}), with
\[
\Phi_i(x_i) = \frac{1}{2} (y_i-x_i)^T\Sigma^{-1}(y_i-x_i),\quad
\Psi_i(r_i) = \lambda \|r_i\|_2.
\]
\newpage
\paragraph*{ADMM steps.}
For this problem, steps (\ref{e-admm-11}) and (\ref{e-admm-12}) of ADMM
can be further simplified. Step (\ref{e-admm-11}) involves minimizing an
unconstrained quadratic function in the variable $x_i$, and can be
written as
\[
x_i^{k+1} = (\Sigma^{-1}+\rho I)^{-1}
(\Sigma^{-1} y_i + \rho(z_i^k-u_i^k)).
\]
Step (\ref{e-admm-12}) is
\[
r_i^{k+1} := \argmin_{r_i} \{\lambda \|r_i\|_2+(\rho/2)\|r_i
- s_i^k + t_i^k\|_2^2\},
\]
which simplifies to
\begin{equation}\label{eq:thresh}
r_i^{k+1} ={\mathcal{S}}_{\lambda/\rho}(s_i^k -t_i^k),
\end{equation}
where $\mathcal{S}_\kappa$ is the vector soft thresholding operator, defined as
\[
{\mathcal{S}}_\kappa({a})=(1-\kappa/\|a\|_2)_+ {a},\quad {\mathcal{S}}_\kappa({0})=0.
\]
Here the notation $(v)_+ = \max\{0, v\}$ denotes the positive part of the vector $v$.
(For details see \cite{DBLP:journals/ftml/BoydPCPE11}.)

\paragraph*{Variations.}
In some problems, we might expect that individual components of $x_t$ will be
piecewise constant, in which case we can instead use the standard Fused Lasso
method. In the standard Fused Lasso method we solve
\[
\begin{array}{ll}
\mbox{minimize} & \sum_{i=1}^N\frac 1 2
(y_i-x_i)^T\Sigma^{-1}(y_i-x_i)+\lambda \sum_{i=1}^{N-1}
\|r_i\|_1 \\
\mbox{subject to} &  r_i=x_{i+1}-x_i,\quad i = 1,\ldots,N,
\end{array}
\]
with variables $x_1,\ldots,x_N$, $r_1,\ldots,r_{N-1}$. The ADMM updates are the
same, except that instead of doing vector soft thresholding for step
(\ref{e-admm-12}), we perform scalar componentwise soft thresholding, \ie,
\[
(r_i^{k+1})_j ={\mathcal{S}}_{\lambda/\rho}((s_i^k -t_i^k)_j),\quad j=1,\ldots,n.
\]

\subsection{$\ell_1$ Variance filtering}
\label{sec:l1var}

Consider a sequence of vector random variables (of dimension $n$)
\[
Y_i\sim {\mathcal{N}}(0,\Sigma_i), \quad i = 1,\ldots,N,
\]
where $\Sigma_i\in\symm^n_+$ is the covariance matrix for $Y_i$ (which
we assume is fixed but unknown).  Given observations of $y_1,\ldots,y_N$, our
goal is to estimate the sequence of covariance matrices
$\Sigma_1,\ldots,\Sigma_N$, under the assumption that it is
piecewise constant, \ie, it is often the case that $\Sigma_{i+1} =
\Sigma_i$. In order to obtain a convex problem, we use the
inverse covariances $X_i = \Sigma_i^{-1}$ as our variables.

The Fused Group Lasso method for this problem involves solving
\[
\begin{array}{ll}
\mbox{minimize} & \sum_{i=1}^N \Tr(X_iy_iy_i^T)-\log\det X_i
+\lambda \sum_{i=1}^{N-1}\|R_i\|_F \\
\mbox{subject to} & R_i = X_{i+1}-X_i, \quad i = 1,\ldots,N-1,
\end{array}
\]
where our variables are $R_i\in\symm^n$, $i = 1,\ldots,N-1$, and
$X_i\in\symm^n_+$, $i = 1,\ldots,N$. Here,
\[
\|R_i\|_F = \sqrt{\Tr(R_i^TR_i)}
\]
is the Frobenius norm of $R_i$. Let $X_1^\star, \ldots,X_N^\star$,
$R_1^\star,\ldots,R_{N-1}^\star$ denote an optimal point, our estimates
of $\Sigma_1,\ldots,\Sigma_N$ are $(X_1^\star)^{-1},\ldots,(X_N^\star)^{-1}$.

\paragraph*{ADMM steps.}
It is easy to see that steps (\ref{e-admm-11}) and (\ref{e-admm-12})
simplify for this problem. Step (\ref{e-admm-11}) requires solving
\[
X_i^{k+1} := \argmin_{X_i\succ 0}  \{
\Phi_i(X_i)
+(\rho/2)\|X_i - Z_i^k + U_i^k\|_2^2\},
\]
where
\[
\Phi_i(X_i) = \Tr(X_iy_iy_i^T)-\log\det X_i.
\]
This update can be solved analytically, as follows.
\begin{enumerate}
\item Compute the eigenvalue decomposition of
\[
\rho\left(Z_i^k
-U_i^k\right)-y_iy_i^T=Q\Lambda Q^T
\]
where $\Lambda={\bf diag}(\lambda_1,\ldots , \lambda_n)$.
\item Now let
\[
\mu_j := \frac{\lambda_j+\sqrt{\lambda_j^2+4\rho}}{2\rho},\quad j = 1,\ldots,n.
\]
\item Finally, we set
\[
X_i^{k+1} = Q \diag(\mu_1,\ldots,\mu_n) Q^T.
\]
\end{enumerate}
For details of this derivation, see Section 6.5 in
\cite{DBLP:journals/ftml/BoydPCPE11}.

Step (\ref{e-admm-12}) is
\[
R_i^{k+1} := \argmin_{R_i}
\{\lambda \|R_i\|_F+(\rho/2)\|R_i - S_i^k + T_i^k\|_2^2\},
\]
which simplifies to
\[
R_i^{k+1} ={\mathcal{S}}_{\lambda/\rho}(S_i^k - T_i^k),
\]
where $\mathcal{S}_\kappa$ is a matrix soft threshold operator, defined as
\[
{\mathcal{S}}_\kappa(A)=(1-\kappa/\|A\|_F)_+ A,\quad
{\mathcal{S}}_\kappa({0})=0.
\]

\paragraph*{Variations.}
As with $\ell_1$ mean filtering, we can replace the Frobenius norm penalty with
a componentwise vector $\ell_1$-norm penalty on $R_i$ to get the problem
\[
\begin{array}{ll}
\mbox{minimize} & \sum_{i=1}^N
\Tr(X_iy_iy_i^T)-\log\det X_i
+ \lambda \sum_{i=1}^{N-1}\|R_i\|_1 \\
\mbox{subject to} & R_i=X_{i+1}-X_i,
\quad i = 1,\ldots,N-1,
\end{array}
\]
with variables $R_1,\ldots,R_{N-1}\in\symm^n$, and
$X_1,\ldots,X_N\in\symm^n_+$, and where
\[
\|R\|_1 = \sum_{j,k} |R_{jk}|.
\]
Again, the ADMM updates are the same, the only difference is that in step
(\ref{e-admm-12}) we replace matrix soft thresholding with a componentwise
soft threshold, \ie,
\[
(R_i^{k+1})_{l,m} = \mathcal{S}_{\lambda/\rho}((S_i^k
-T_i^k)_{l,m}),
\]
for $l = 1,\ldots,n$, $m = 1,\ldots,n$.

\subsection{$\ell_1$ Mean and variance filtering}
\label{sec:l1mean_var}

Consider a sequence of vector random variables
\[
Y_i\sim {\mathcal{N}}(\bar y_i, \Sigma_i),
\quad i = 1,\ldots,N,
\]
where $\bar y_i\in\reals^n$ is the mean, and $\Sigma_i\in\symm^n_+$ is the
covariance matrix for $Y_i$. We assume that the mean and covariance matrix of
the process is unknown.  Given observations $y_1,\ldots,y_N$, our goal is to
estimate the mean and the sequence of covariance matrices
$\Sigma_1,\ldots,\Sigma_N$, under the assumption that they are piecewise
constant, \ie, it is often the case that  $\bar y_{i+1} = \bar y_i$ and
$\Sigma_{i+1} = \Sigma_i$. To obtain a convex optimization problem, we use
 \[
X_i=-\frac{1}{2}\Sigma_t^{-1},\quad m_i=\Sigma_t^{-1}x_i,
 \]
 as our variables. In the Fused Group Lasso method, we obtain our estimates by solving
 \[
 \begin{array}{ll}
\mbox{minimize} & \sum_{i=1}^N -(1/2)\log\det(-X_i)-\Tr(X_iy_iy_i^T)\\
 & \quad\qquad - m_i^T y_i -(1/4)\Tr(X^{-1}_im_im_i^T)\\
 & \quad\qquad + \lambda_1 \sum_{i=1}^{N-1} \|r_i\|_2+\lambda_2 \sum_{i=1}^{N-1} \|R_i\|_F\\
 \mbox{subject to} &  r_i=m_{i+1}-m_i, \quad i = 1,\ldots,N-1, \\
&  R_i=X_{i+1}-X_i, \quad i = 1,\ldots,N-1,
 \end{array}
 \]
with variables $r_1,\ldots,r_{N-1} \in \reals^n$, $m_1,\ldots,m_{N}\in \reals^n$, $R_1,\ldots,R_{N-1}\in\symm^n$, and
$X_1,\ldots,X_N\in\symm^n_+$.

\paragraph*{ADMM steps.}
This problem is also in the form (\ref{e-our-problem}), however, as far as we
are aware, there is no analytical formula for steps  (\ref{e-admm-11}) and
(\ref{e-admm-12}). To carry out these updates, we must solve semidefinite
programs (SDPs), for which there are a number of efficient and reliable
software packages (\cite{TTT:99,Stu:99}).

\section{Numerical Example}
\label{sec:num}
\begin{figure}[ht]
\begin{center}
\includegraphics[width = \columnwidth]{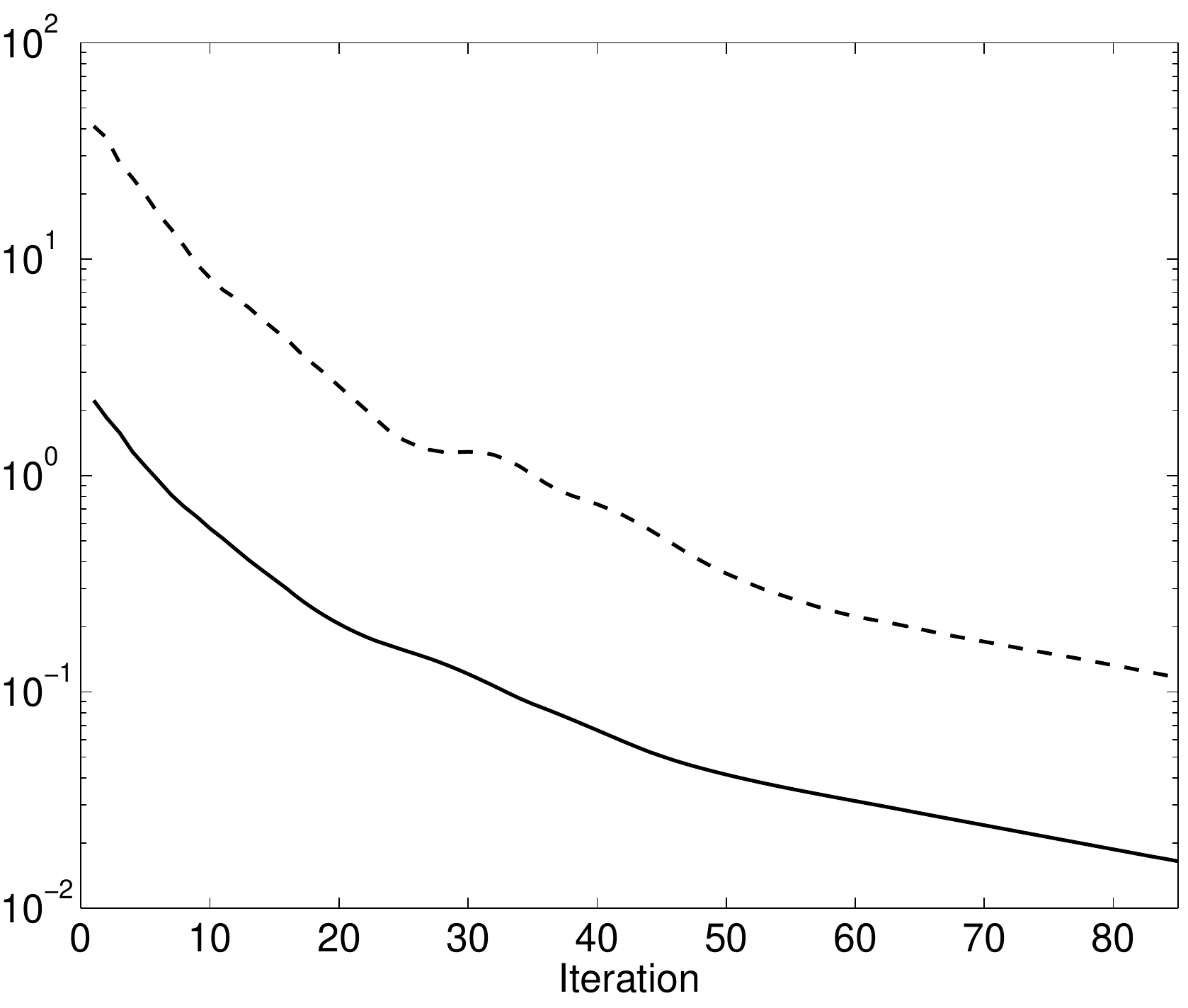}
\caption{Residual convergence: Primal residual $e_p$ (solid line), and dual residual $e_d$
(dashed line).}\label{ex1}\end{center}
\end{figure}

\begin{figure}[ht]
\begin{center}
\includegraphics[width = \columnwidth]{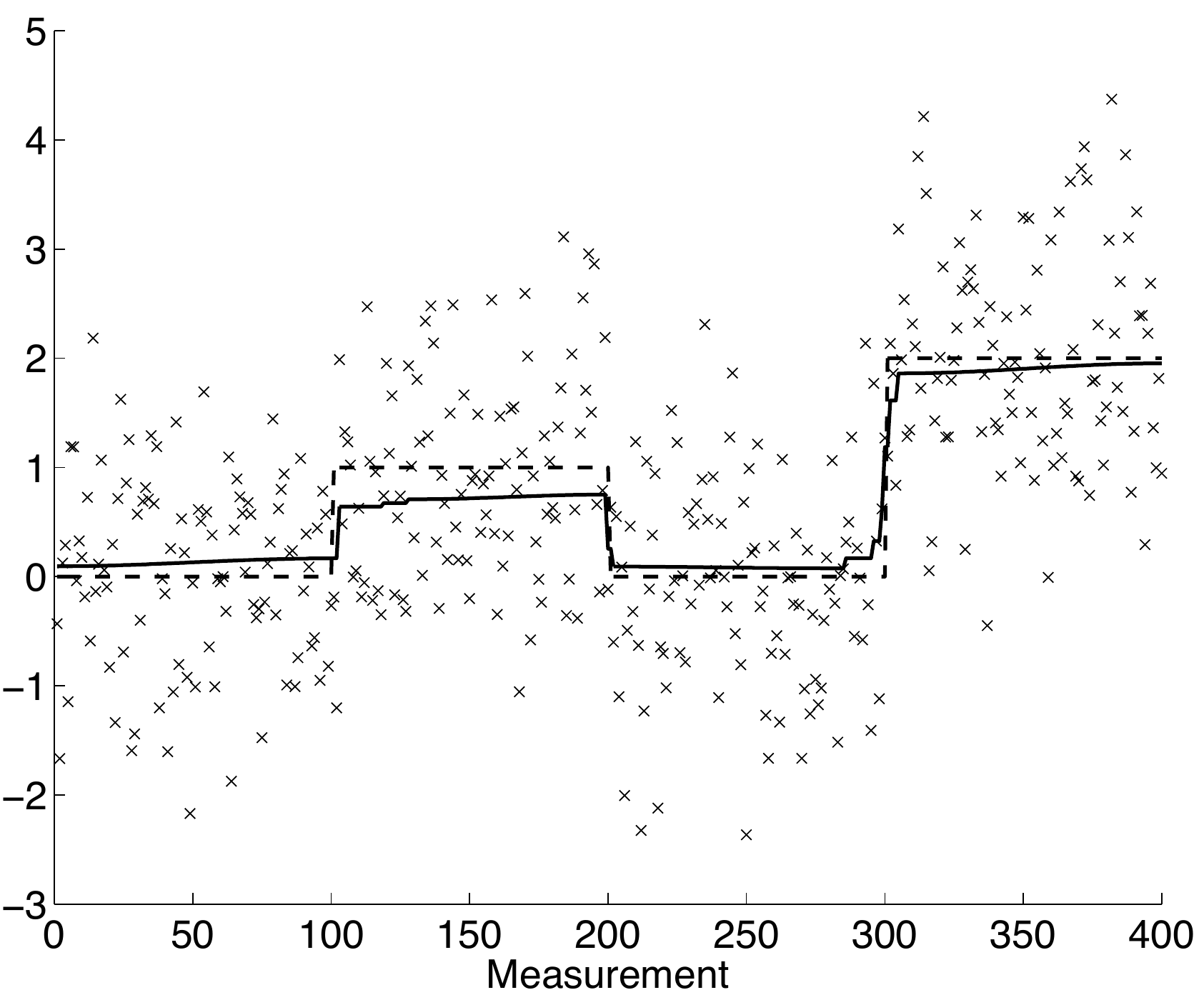}
\caption{Estimated means (solid line), true means (dashed line) and
measurements (crosses).}\label{ex2}\end{center}
\end{figure}

In this section we solve an instance of $\ell_1$ mean filtering with $n = 1$,
$\Sigma = 1$, and $N = 400$, using the standard Fused Lasso method.  To improve
convergence of the ADMM algorithm, we use over-relaxation with $\alpha=1.8$,
see \cite{DBLP:journals/ftml/BoydPCPE11}. The parameter $\lambda$ is chosen as
approximately 10\% of $\lambda_\mathrm{max}$, where $\lambda_\mathrm{max}$ is
the largest value that results in a non-constant mean estimate. Here,
$\lambda_\mathrm{max} \approx 108$ and so $\lambda=10$.  We use an absolute
plus relative error stopping criterion, with $\epsilon^\mathrm{abs} = 10^{-4}$
and $\epsilon^\mathrm{rel} = 10^{-3}$. Figure \ref{ex1} shows convergence of the
primal and dual residuals. The resulting estimates of the means
are shown in Figure~\ref{ex2}.

We solved the same $\ell_1$ mean filtering problem using CVX, a package for
specifying and solving convex optimization problems (\cite{cvxProg}).  CVX
calls generic SDP solvers SeDuMi (\cite{TTT:99}) or SDPT3 (\cite{Stu:99}) to
solve the problem.  While these solvers are reliable for wide classes of
optimization problems, and exploit sparsity in the problem formulation,
they are not customized for particular problem families, such as ours.
The computation time for CVX is approximately 20 seconds. Our
ADMM algorithm (implemented in C), took $2.2$ \textit{milliseconds} to produce
the same estimates.  Thus, our algorithm is approximately 10000 times faster
compared with generic optimization packages. Indeed, our implementation
does \textit{not} exploit the fact that steps 1 and 3 of ADMM can be
implemented independently in parallel for each measurement. Parallelizing
steps 1 and 3 of the computation can lead to further speedups.  For example,
simple multi-threading on a quad-core CPU would result in a
further $4 \times$ speed-up.


\section{Conclusions}
\label{sec:con}

In this paper we derived an efficient and scalable method for an optimization
problem (\ref{e-our-problem}) that has a variety of applications in control and
estimation. Our custom method exploits the structure of the problem via a
distributed optimization framework. In many applications, each step of the
method is a simple update that typically involves solving a set of linear
equations, matrix multiplication, or thresholding, for which there are
exceedingly efficient libraries. In numerical examples we have shown that we
can solve problems such as $\ell_1$ mean and variance filtering many orders of
magnitude faster than generic optimization solvers such as SeDuMi or SDPT3.

The only tuning parameter for our method is the regularization parameter
$\rho$. Finding an optimal $\rho$ is not a straightforward problem, but
\cite{DBLP:journals/ftml/BoydPCPE11} contains many heuristics that work
well in practice. For the $\ell_1$ mean filtering example, we find that
setting $\rho \approx \lambda$ works well, but we do not have a formal
justification.




\bibliography{refs_cristian,refs_bo}

\end{document}